\documentclass[sort&compress, numafflabel]{elsarticle}

\usepackage[]{natbib}
\usepackage{graphicx} 
\usepackage{subfiles}
\usepackage{multirow}
\usepackage{geometry}
\usepackage{url}
\usepackage{hyperref}
\usepackage{booktabs}
\usepackage{verbatim}
\usepackage{amsmath}
\usepackage{booktabs}
\usepackage{tabularx}
\usepackage{array}
\usepackage[table]{xcolor}
\usepackage{soul}            
\usepackage{array}    
\usepackage{graphicx} 
\usepackage{ragged2e}
\usepackage{changepage}
\usepackage{adjustbox}
\usepackage{colortbl}
\usepackage{xcolor}
\usepackage{float}
\usepackage{longtable}
\soulregister{\textbf}{1}  
\newcommand{\hlred}[1]{\sethlcolor{red!30}\hl{#1}\sethlcolor{yellow}}
\newcommand{\hlgreen}[1]{\sethlcolor{green!30}\hl{#1}\sethlcolor{yellow}}
\newcolumntype{Y}{>{\RaggedRight\arraybackslash}X}
\newcolumntype{P}[1]{>{\RaggedRight\arraybackslash}p{#1}}

\newif\ifsubfile
\subfiletrue

\title{A Systematic Analysis of Large Language Models with RAG-enabled Dynamic Prompting for Medical Error Detection and Correction}

\author{Farzad Ahmed\textsuperscript{1}, Joniel Augustine Jerome\textsuperscript{1}, Meliha Yetisgen\textsuperscript{2}, Özlem Uzuner\textsuperscript{1}}

\address{\textsuperscript{1}George Mason University, \textsuperscript{2}University of Washington}

\begin{document}
\begin{abstract}

\textit{Objective:} 
Clinical documentation is prone to factual, diagnostic, and management errors that can compromise patient safety. Large language models (LLMs) offer potential for automatic error detection and correction, but their behavior under different prompting strategies remains poorly understood. This study systematically analyzes the strengths and weaknesses of zero-shot prompting, static prompting with random exemplars (SPR), and retrieval-augmented generation (RAG)-enabled dynamic prompting (RDP), to address medical error detection and correction through three subtasks: \textit{error flag detection}, \textit{error sentence detection}, and \textit{error correction}.

\textit{Methods:} 
Using the MEDEC dataset, we evaluated nine instruction-tuned LLMs—including GPT, Claude, Gemini and OpenAI's o-series variants—across the three subtasks. This diverse set of models, spanning compact and frontier-scale architectures, allowed us to assess how model size, architecture, and training strategies influence performance in medical error detection and correction. Performance on error flag and sentence detection was measured with Accuracy and Recall, False-Positive Rate (FPR) measured performance on error flag detection, and Aggregate Score (AggScore) of ROUGE-1, BLEURT, and BERTScore measured performance on error correction. We analyzed LLM outputs, examining representative cases, failure patterns, and LLM–clinician differences to explain quantitative trends. 

\textit{Results:} 
Zero-shot prompting exhibited low recall in error flag detection and error sentence detection, often missing atypical or abbreviation-heavy errors. SPR improved recall in both detection
tasks but increased FPR. Across all nine LLMs, these trends remained consistent—RDP
that retrieves semantically relevant exemplars during inference, reduced FPR by about 15\% in error
flag detection, increased recall by 5-10\% in error sentence detection, and improved the contextual
and semantic accuracy of error correction.

\textit{Conclusion:} 
Our systematic analysis of nine LLMs with three prompting strategies highlights limitations of zero-shot and SPR for medical error detection and correction. RDP outperforms zero-shot and SPR across all LLMs, grounding predictions in relevant exemplars for effective medical error detection and correction.

\end{abstract}


\maketitle

\subfilefalse

\section{Introduction}

Clinical documentation is a cornerstone of patient care but can contain medical errors that can compromise safety, delay treatment, and propagate across systems. These errors can arise from diagnostic inaccuracies, transcription inconsistencies, or management decisions, and they remain a leading contributor to preventable adverse events in healthcare \cite{makary2016medical,rajkomar2019machine,miotto2018deep}. Manual review of clinical notes is resource-intensive and infeasible at scale, motivating the development of automated methods for reliable error detection and correction.

Natural Language Processing (NLP) has played a central role in structuring and analyzing unstructured clinical text, improving efficiency in clinical reasoning, information retrieval, and patient outcome prediction \cite{rajkomar2019machine,miotto2018deep,johnson2016mimic}. The emergence of large language models (LLMs) has accelerated progress in biomedical NLP, combining contextual understanding with reasoning capabilities. Domain-specific models such as BioBERT \cite{lee2020biobert} and PubMedGPT \cite{pubmedgpt2022}, alongside general-purpose architectures like PaLM \cite{chowdhery2023palm} and Med-PaLM \cite{singhal2023large}, have demonstrated strong performance on benchmark biomedical tasks. However, their ability to identify and correct medical errors remains underexplored.

The MEDIQA-CORR shared task \cite{abacha2024overview} and the MEDEC dataset \cite{benabacha2025medec} offer a standardized benchmark for evaluating automatic medical error detection and correction. MEDEC defines three subtasks: (1) \textit{Error flag detection}—determining whether a clinical note contains an error; (2) \textit{Error sentence detection}—identifying the sentence containing erroneous information; and (3) \textit{Error correction}—producing a corrected version of that sentence. The dataset includes five error types—\textit{diagnosis}, \textit{causal organism}, \textit{management}, \textit{treatment}, and \textit{pharmacotherapy}.

While biomedical LLMs have achieved promising results \cite{singhal2023large, thirunavukarasu2023large, ji2023survey}, most prior systems rely on static prompting \cite{gundabathula2024promptmind, benabacha2025medec}, i.e., prompts with randomly selected exemplars that are applied to all inputs, or classification-only architectures, limiting generalization across institutions and documentation styles. Furthermore, these systems tend to show low recall for atypical errors, high false positive rates (FPR) due to exemplar mismatch, and difficulties handling abbreviations and shorthand \cite{ji2023survey,thirunavukarasu2023large, gundabathula2024promptmind}. 

Retrieval-Augmented Generation (RAG) \cite{lewis2020retrieval} dynamically retrieves semantically relevant clinical exemplars tailored to each input during inference, allowing for dynamic prompting that can mitigate errors due to exemplar mismatch. The retrieval is dynamic because the exemplars in the prompt are selected for each input sample specifically and change from input sample to input sample.  We refer to the resulting approach as \textit{RAG-enabled dynamic prompting} (RDP).

\textbf{Contributions.}  
In this paper:
\begin{itemize}
\item We introduce RDP, enabling exemplar selection based on semantic similarity to each input sample rather than random sampling. 
\item We conduct the first systematic analysis of zero-shot, static prompting with random exemplars (SPR), and RDP for medical error detection and correction using the MEDEC benchmark.  
\item We evaluate nine instruction-tuned LLMs—including GPT-4o, GPT-4o-mini, GPT-4.1, GPT-4.1-mini, GPT-5, o1-mini, o4-mini, Claude 3.5 Sonnet, and Gemini 2.0 Flash—across all three MEDIQA-CORR subtasks using three prompting strategies.
\item We systematically analyze how model size, architecture, and prompting strategies impact performance in medical error detection and correction across nine LLMs.
\item We complement quantitative evaluation with a qualitative analysis of representative model outputs, identifying error patterns, failure modes, and clinician--model discrepancies to explain observed trends.  
\end{itemize}

Together, these analyses provide a comprehensive view of how RDP mitigates the weaknesses of SPR, enhancing recall, reducing FPR, and improving LLM-based error detection and correction. Specifically, we find that \textbf{RDP} (1) improves recall in \textit{error sentence detection} by retrieving semantically aligned exemplars that better match each clinical input; (2) reduces FPR in \textit{error flag detection} by exposing models to correct, in-domain examples that discourage unnecessary corrections; and (3) enhances handling of abbreviations and shorthand by retrieving cases with similar linguistic patterns, enabling more accurate interpretation of clinical shorthand and numeric expressions. 

\section*{Statement of Significance}

\noindent\begin{tabularx}{\textwidth}{@{}p{0.10\linewidth}p{0.90\linewidth}@{}}
\hline
\textbf{Problem} & Clinical documentation can contain medical errors that, if undetected, can compromise patient safety. Manual review is resource-intensive, highlighting the need for automatic systems that can flag errors, identify the erroneous sentences, and generate corrected versions.\\ \hline
\textbf{What is Already Known} & Large language models (LLMs) have shown promise in biomedical NLP but have had limited success in \textit{error flag detection}, \textit{error sentence detection}, and \textit{error correction}: they tend to achieve low recall for atypical errors, over-generate unnecessary corrections for already correct sentences, and face difficulties handling abbreviations and shorthand notations. Most prior work evaluates classification or detection of medical errors in isolation, with limited exploration of RDP for error detection and correction. \\ \hline
\textbf{What This Paper Adds} & This paper systematically evaluates SPR and RDP strategies for medical error flag detection, error sentence detection, and error correction. It demonstrates that RDP improves (1) recall of true error sentence detection, (2) reduces FPR in error flag detection, (3) enhances interpretation of abbreviations and shorthand, and (4) improves contextual and semantic accuracy of error correction. RDP grounds LLM predictions in clinically relevant examples, achieving state-of-the-art performance across all three subtasks. \\ \hline
\textbf{Who Would Benefit} & Automatic error detection and correction can support clinicians, clinical researchers, informatics practitioners, as well as patients. Our approach can lead the way in integrating dynamic exemplars in prompting strategies, to improve the relevance of exemplars to the input and ultimately improve performance.  \\ \hline
\end{tabularx}

\ifsubfile
\bibliography{mybib}
\fi

\section{Related Work}

Prior work on medical error detection and correction can be grouped into: benchmarks and datasets, shared task systems, and computational strategies.  

\subsection*{Benchmarks and Datasets}
\textbf{MEDEC corpus} \cite{benabacha2025medec} is designed to evaluate LLMs on medical error detection and correction. MEDEC provides thousands of annotated clinical texts covering multiple error types, including diagnosis errors, causal organism, management, treatment, and pharmacotherapy errors. While MEDEC is the primary dataset for this study (see Section~\ref{sec:dataset}), several other resources have contributed to the development of evaluation standards for medically inaccurate information. For example, the \textbf{Spanish Real-Word Error Corpus} \cite{bravocandel2020spanish} and the \textbf{MEDIC} dataset of medication directions \cite{pais2024medic} address medical error correction in multilingual and pharmacy settings. Misinformation-oriented datasets such as \textbf{COVID-Lies} \cite{hossain2020covidlies}, \textbf{ReCOVery} \cite{zhou2020recovery}, and \textbf{CoAID} \cite{cui2020coaid} focus on health misinformation in social media, while fact-checking benchmarks like \textbf{SciFact} \cite{wadden2020scifact}, \textbf{HealthVer} \cite{sarrouti2021healthver}, \textbf{PubHealth} \cite{kotonya2020explainable}, and \textbf{BEAR-FACT} \cite{wuhrl2023bearfact} provide evidence-based claim verification. Recent efforts also target model hallucination in medical reasoning using evaluation suites such as \textbf{Med-HALT} \cite{pal2024medhalt}. Together, these datasets reflect a growing ecosystem of benchmarks that support research on medical error detection, misinformation mitigation, and fact checking.

\subsection*{Shared Task Systems}  
The \textbf{MEDIQA-CORR 2024 shared task} \cite{benabacha2024mediqa} divided medical error detection and correction into three subtasks: error flag detection, error sentence detection, and error correction. The official \textbf{MEDEC benchmark evaluation} \cite{benabacha2025medec} showed that instruction-tuned LLMs such as Claude 3.5 and GPT-4o reached an accuracy of 0.70 in error flag detection and 0.66 in error sentence detection, while medical doctors continue to outperform models on error correction, with the best medical doctor result of 0.7742 compared to the best model result of 0.7043. Interpretable strategies to the task included PromptMind, which combined chain-of-thought prompting with an ensemble of LLMs \cite{gundabathula2024promptmind}, and HSE NLP which integrated biomedical entity recognition with MeSH-based graph reasoning \cite{valiev2024hse}. PromptMind reached an accuracy of 0.6216 in error flag detection, 0.6086 in error sentence detection, and an AggScore of 0.7866 in error correction (ROUGE-1 = 0.8070, BLEURT = 0.7470). HSE NLP obtained 0.5222 and 0.5200 in error flag and sentence detection, respectively, with an AggScore in error correction of 0.7806 (ROUGE-1 = 0.7795, BLEURT = 0.7564). Nonetheless, current results show that accurate error detection and correction remains an open challenge.

\subsection*{Computational Strategies}  
Recent work has examined how computational paradigms influence clinical reasoning performance. Cai et al. \cite{cai2025train} contrasted \textbf{train-time computation models}, which concentrate reasoning ability in extensive pre-training and require relatively little computation at inference (e.g., GPT-4), with \textbf{test-time computation models}, which allocate substantial computation during inference through extended reasoning traces or search (e.g., GPT o1, DeepSeek R1). Train-time models showed strength in error detection, while test-time models excelled at correction. A hybrid logistic regression ensemble achieved the best balance, illustrating complementary advantages across approaches.

\subsection*{Persistent Gaps}  
Taken together, benchmarks, shared task systems, and computational strategies highlighted two major gaps. First, existing studies emphasize quantitative metrics but rarely examine \emph{how} LLMs fail across different error types. As a result, recurring weaknesses remain underexplored—such as low recall in error flag and sentence detection, elevated FPR that triggers unnecessary corrections of valid sentences, and limited handling of clinical abbreviations and shorthand. Second, while shared task systems have experimented informally with retrieval, the potential of RAG to dynamically adapt prompting has not been systematically evaluated. In short, prior work has not yet established how prompting strategies influence model behavior or how retrieval-based methods might mitigate specific error patterns.

\subsection*{Our Contribution}  
We address these problems by performing the first systematic analysis of \emph{prompting} strategies---zero-shot, SPR, and RDP---for medical error flag detection, error sentence detection and error correction. By retrieving semantically aligned exemplars for each input, RDP improves recall in error sentence detection, reduces false-positive corrections of accurate sentences, and enhances reasoning over clinical shorthand. Our results show that dynamic prompting reliably bridges the gap between high-level error detection and fine-grained clinical correction.

\ifsubfile
\bibliography{mybib}
\fi

\section{Materials and Methods}

\subsection{Dataset}
\label{sec:dataset}

We use the \textbf{MEDEC} corpus \cite{benabacha2025medec}, released as part of the \textbf{MEDIQA-CORR 2024 shared task} \cite{abacha2024overview}. MEDEC consists of two subsets: the \textbf{MS collection}, derived from MedQA-style board examination scenarios \cite{jin2021disease}, and the \textbf{UW collection}, derived from de-identified clinical notes. Together, they provide 3,848 clinical texts, each either correct or containing a single medically plausible error manually injected by clinically trained annotators to preserve contextual coherence.

The errors span five types—\textit{diagnosis}, \textit{causal organism}, \textit{management}, \textit{treatment}, and \textit{pharmacotherapy}.  As an example from the MEDEC dataset, the sentence \textit{“Patient is diagnosed with \textbf{aortic stenosis} after physical examination reveals a double apical impulse.”}  contains an error and is corrected as
\textit{“Patient is diagnosed with \textbf{hypertrophic cardiomyopathy} after physical examination reveals a double apical impulse.”} 

Table~\ref{tab:dataset-splits} presents dataset splits, as well as prevalence of errors in each of the splits. MEDEC dataset is publicly available at: \url{https://github.com/abachaa/MEDEC}.   

\begin{table}[h]
\centering
\small
\begin{tabular}{lcccc}
\hline
\textbf{Collection} & \textbf{Training} & \textbf{Validation} & \textbf{Test} & \textbf{Total} \\
\hline
MS   & 2,189 & 574 & 597 & 3,360 \\
UW   &   --  & 160 & 328 &   488 \\
\textbf{MEDEC} & \textbf{2,189} & \textbf{734} & \textbf{925} & \textbf{3,848} \\
\hline
\# texts without errors & 970 (44.3\%) & 335 (45.6\%) & 450 (48.7\%) & 1,755 (45.6\%) \\
\# texts with errors    & 1,219 (55.7\%) & 399 (54.4\%) & 475 (51.3\%) & 2,093 (54.4\%) \\
\hline
\end{tabular}
\caption{MEDEC dataset \cite{benabacha2025medec}.}
\label{tab:dataset-splits}
\end{table}

Across the MEDEC dataset, \textbf{management errors} constitute roughly 45–50\% of all annotated cases, followed by \textbf{diagnosis errors} at around 30–35\%. \textbf{Treatment} and \textbf{pharmacotherapy errors} together account for approximately 15–20\%, while \textbf{causal organism errors} are the least frequent (below 5\%). This distribution remains consistent across the MS and UW subsets and their respective training, validation, and test splits.

\subsection{Task Formulation}
MEDIQA-CORR shared task contained three subtasks:  

\begin{itemize}
    \item \textbf{Subtask A: error flag detection.} Predict whether a clinical note contains an error.
    \item \textbf{Subtask B: error sentence detection.} Identify the sentence containing the error.
    \item \textbf{Subtask C: error correction.} Generate a corrected version of the erroneous sentence.
\end{itemize}

\subsection{Language Models}
We evaluated nine recent LLMs, covering both compact variants such as \texttt{o1-mini}, \texttt{o4-mini}, \texttt{GPT-4o-mini}, \texttt{GPT-4.1-mini}, and \texttt{Claude 3.5 Sonnet}, optimized for efficiency and frontier-scale systems such as \texttt{GPT-4o}, \texttt{GPT-4.1}, \texttt{GPT-5}, and \texttt{Gemini 2.0 Flash}, optimized for reasoning. This diversity allows us to evaluate how model size, architecture, and training strategies impact performance in medical error detection and correction across different types of LLMs.

\paragraph{GPT-4o and GPT-4o-mini}
\texttt{GPT-4o} (OpenAI, 2024) is a multimodal LLM providing GPT-4–level intelligence with improved efficiency and latency \cite{openai2024gpt4o}. Its smaller sibling, \texttt{GPT-4o-mini} ($\sim$8B parameters), is optimized for instruction-following and lightweight reasoning while retaining strong alignment capabilities \cite{openai2024gpt4omini}.

\paragraph{GPT-4.1 and GPT-4.1-mini}
\texttt{GPT-4.1} (OpenAI, late 2024) introduces refinements in reasoning robustness and factual grounding, improving multi-turn consistency and reliability \cite{openai2024gpt4.1}. We additionally evaluate \texttt{GPT-4.1-mini}, a compact version intended to balance efficiency with reasoning performance, particularly suitable for controlled few-shot settings \cite{openai2024gpt4.1mini}.

\paragraph{GPT-5}
\texttt{GPT-5} (OpenAI, 2025) represents the latest generation of frontier-scale models, trained for improved factual accuracy, consistency, and domain adaptability. GPT-5 has shown superior performance across open-domain reasoning tasks and serves as a strong benchmark \cite{openai2025gpt5}.

\paragraph{o1-mini and o4-mini}
The \texttt{o1-mini} and \texttt{o4-mini} families (OpenAI, 2024) are smaller-scale optimized reasoning models, reported at $\sim$100B and $\sim$8B parameters, respectively. These models were designed for fast inference while maintaining high reasoning quality, and represent OpenAI’s more compact but capable architectures \cite{openai2024o1mini}, \cite{openai2024o4mini}.

\paragraph{Claude 3.5 Sonnet}
\texttt{Claude 3.5 Sonnet} (Anthropic, October 2024) is a $\sim$175B parameter model trained with Constitutional AI for alignment and factual reliability. Claude models are designed to reduce hallucination rates and improve transparency \cite{anthropic2024claude}.

\paragraph{Gemini 2.0 Flash}
\texttt{Gemini 2.0 Flash} (Google DeepMind, 2024) is the most recent release in the Gemini series. It is optimized for high efficiency and fast reasoning, with strong multilingual and domain-adaptation capabilities. While precise parameter counts remain undisclosed, Gemini 2.0 Flash represents Google’s current state-of-the-art foundation model family \cite{gemini}.

\vspace{0.5em}
In addition to the above LLMs, we experimented with ChatGPT (gpt-4) \cite{openai2023chatgpt}, Phi-3 \cite{microsoft2024phi}, BioGPT \cite{luo2022biogpt}, and open-source models from the LLaMA family \cite{touvron2023llama,touvron2023llama2,meta2024llama3}, but omit those experiments from this paper because they underperformed substantially relative to the instruction-tuned models listed above. Together, our selected nine models span diverse architectures, parameter scales, and training strategies, enabling a comprehensive evaluation of how model design impacts performance in error flag detection, error sentence detection, and error correction.

\subsection{Approaches}
We implemented an LLM-based framework capable of performing all subtasks in an end-to-end manner. Three prompting strategies were compared: zero-shot, static prompting with random exemplars (SPR), RAG-enabled dynamic prompting (RDP). 

\paragraph{Zero-Shot Prompting}
A baseline in which the model is presented only with task instructions and the clinical narrative, without any exemplars. The zero-shot prompt was adapted directly from the MEDEC benchmark paper \cite{benabacha2025medec} to ensure comparability with prior work, and further refined on the validation set to confirm clarity and alignment with the three subtasks. All prompting strategies employed a standardized instruction template to ensure consistency across models. The complete prompt is provided in \ref{app:prompt-template}. 

\begin{figure}[h]
    \centering
    \includegraphics[width=0.7\linewidth]{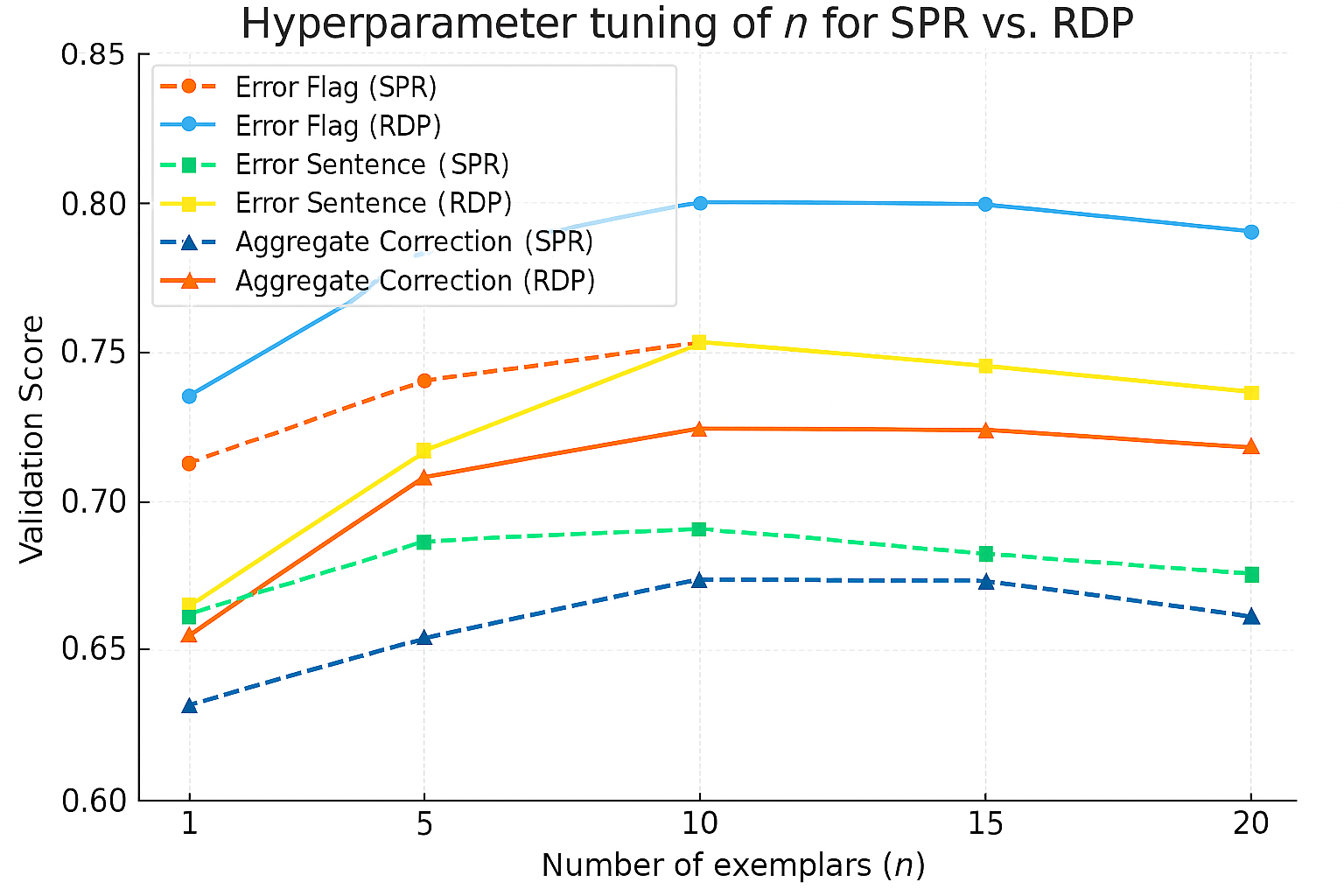}
    \caption{Validation performance of GPT-4.1 vs number of exemplars ($n$) 
    for SPR (dashed) and RDP (solid). 
    Error Flag Detection, Error Sentence Detection, and Error Correction scores 
    all peak at $n=10$.}
    \label{fig:hyperparam_tuning}
\end{figure}

\paragraph{Static Prompting with Random Exemplars (SPR)}  
We evaluated static $n$-shot prompting with random exemplars as baselines. In the static $n$-shot setting, the model is presented with the clinical narrative and $n$ \emph{randomly sampled exemplars} from the training set, appended to the same instructions used in zero-shot prompting. Static $n$-shot baselines mirror prior shared-task systems \cite{benabacha2025medec} while allowing us to probe whether varying the number of random exemplars provides benefits. To decide on the value of $n$, we carried out parameter tuning on the validation set (Figure~\ref{fig:hyperparam_tuning}).  GPT-4.1 achieved the strongest overall validation performance across models and thus served as the representative system for selecting $n$. $n=10$ gave the best overall performance across subtasks and was adopted for all $n$-shot experiments.

\paragraph{Retrieval-Augmented Generation (RAG)-enabled Dynamic Prompting (RDP)}  
RDP dynamically selects semantically relevant exemplars per input clinical note from the training set which we embedded into a Chroma vector database using OpenAI’s \texttt{text-embedding-3-large} model accessed via LangChain \cite{openai2024embeddings, chroma2024,garg2024langchain}. At inference time, the $n$ most semantically similar training cases to the input clinical note are retrieved and appended to the instructions used in zero-shot prompting. The value of $n$ was fixed to 10, consistent with the tuning described in the SPR section, to ensure a fair comparison. We used cosine similarity with input clinical note for dynamic exemplar retrieval after comparing with dot product on the validation set.  We identified \texttt{text-embedding-3-large} as the best embedding backbone for this purpose after evaluating \texttt{BioClinicalBERT} \cite{alsentzer2019publicly}, \texttt{SapBERT (PubMedBERT-fulltext)} \cite{liu2021sapbert}, and \texttt{all-mpnet-base-v2} \cite{reimers2019sentence}, which showed comparatively lower overall AggScores. 

To support efficient exemplar retrieval, we compared multiple vector database backends. 
FAISS \cite{johnson2019billion} offered high-speed similarity search and scalable indexing but required more engineering effort for integration. 
Weaviate \cite{weaviate2024} provided flexible metadata handling and hybrid search capabilities, though at the cost of slower runtime in our setting. 
Chroma \cite{chroma2024}, while simpler in scope, integrated seamlessly with LangChain and produced consistent nearest-neighbor retrieval across repeated trials.  

\subsection{Proposed RAG-enabled Dynamic Prompting (RDP)}
Our RDP approach consists of two major steps: Vector store construction and RAG-enabled inference.
\subsubsection{Vector Store Construction}
\textbf{Preprocessing.} We normalized missing values (e.g., \texttt{NA}), cast sentence IDs to strings, and preserved expert corrections. Each record was structured into a document containing segmented sentences, the annotated error sentence ID, and the gold-standard correction.  

\textbf{Chunking and Embedding.} Documents were split into semantically coherent chunks using LangChain’s \texttt{RecursiveCharacterTextSplitter} \cite{garg2024langchain} to respect token limits. Each chunk was embedded using \texttt{text-embedding-3-large} and stored in a Chroma vector database \cite{chroma2024} managed via LangChain.

\subsubsection{RAG-Based Inference}
\begin{figure}[h]
    \centering
    \includegraphics[width=0.60\textwidth,trim={7.7cm 4.5cm 9.5cm 4.5cm},clip]{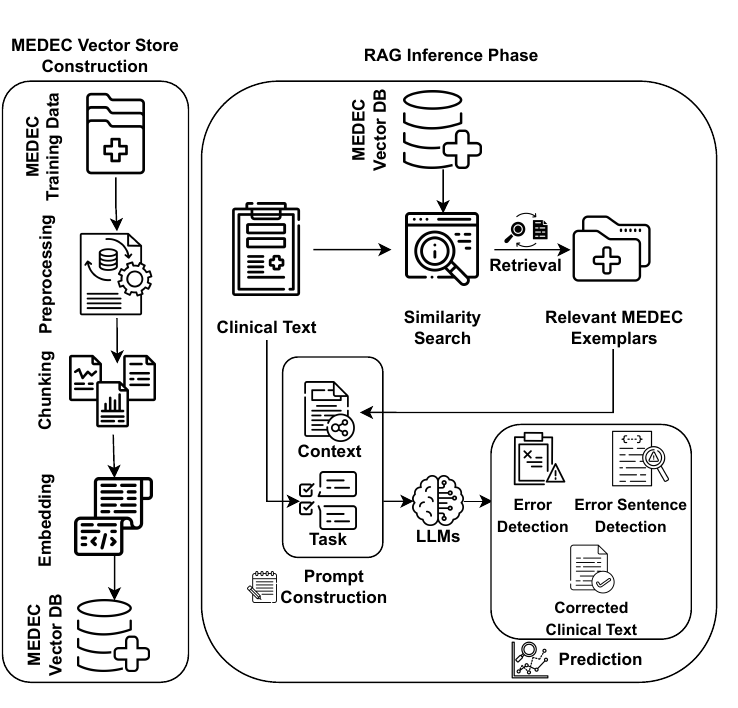}
    \caption{Proposed RDP framework.}
    \label{fig:proposed_method}
\end{figure}

During inference (see Figure~\ref{fig:proposed_method}), 
(1) The input text is embedded and matched to the $n{=}10$ nearest neighbors in the vector store using cosine similarity. The vector store was constructed exclusively from the training split of the MEDEC corpus, ensuring strict separation from test set.  
(2) Retrieved cases are concatenated to the task instructions from zero-shot prompting. 

\subsection{Evaluation Metrics}
To assess performance on the three MEDIQA-CORR tasks we used standard evaluation metrics. 
For error flag detection and error sentence detection, we report \textbf{Accuracy}, which measures the proportion of correctly predicted labels or sentence indices over the total test set.

For a more granular understanding of model behavior across error types, we additionally 
report \textbf{Recall}, computed on the subset of test examples where an error was present 
(i.e., error flag = 1), grouped by error type.

To evaluate error correction, we used a combination of lexical and semantic 
similarity metrics shown to correlate well with expert judgments on clinical text correction 
\citep{abacha2023mediqa}. These include \textbf{ROUGE-1} \citep{lin2004rouge}, \textbf{BERTScore} 
(using \texttt{microsoft/deberta-xlarge-mnli}) \citep{zhang2019bertscore}, and \textbf{BLEURT} 
\citep{sellam2020bleurt}. We also report the \textbf{Aggregate Score (AggScore)}, calculated 
as the arithmetic mean of ROUGE-1, BERTScore, and BLEURT, providing an overall indicator of 
correction quality. We computed error correction scores when both the reference and 
system corrections were available (i.e., not NA).

Finally, we measured the \textbf{False Positive Rate (FPR)} for the best-performing model. FPR quantifies how often models incorrectly flag errors in 
sentences that are in fact correct.

\subsection{Implementation Details}
All experiments were conducted in Python 3.10. RAG-enabled dynamic prompts were constructed with a maximum 
context window of 128k tokens where supported. Random seeds were fixed across runs to ensure reproducibility. Experiments were executed on a secure server with NVIDIA 
A100 GPUs (80GB memory) for vector database operations and embedding generation, while model 
inference was API-based and did not require local fine-tuning. On average, retrieval-augmented 
inference required 1.6 seconds per case and increased token usage by 11\% compared to SPR ten-shot.  Full evaluation required approximately 60 GPU hours for embedding generation and $\sim$\$100 in API usage.

We released the full code base on GitHub\footnote{Link to code: \url{https://github.com/Farzad-1996/MedicalError}}, including (i) the RDP pipeline and implementation details, (ii) prompt templates, and (iii) evaluation scripts.

\subsection{Comparison with Physicians}
To contextualize model performance, we reference the physician annotations reported in the original 
MEDEC benchmark paper \cite{benabacha2025medec}. In that study, two practicing physicians 
independently reviewed 569 clinical notes from the 925-note test set, with 242 cases double-annotated 
to estimate inter-annotator agreement (IAA). Agreement between the 
annotators was moderate, with 69.0\% accuracy for error flag detection and 57.9\% for error sentence 
detection; highlighting the inherent difficulty of the task, even for experts.

\subsection{Ethics and Data Governance}
The UW subset consists of de-identified notes and was used under a data usage agreement. Data governance followed the MEDIQA-CORR guidelines, and all experiments adhered to FAIR principles for reproducibility.

\ifsubfile
\bibliography{mybib}
\fi

\section{Results}

\subsection{Comparison of Zero-shot, SPR, and RDP}

To disentangle the role of retrieval quality from exemplar count, we evaluated five prompting configurations:
(i) \textbf{Zero-shot} (no exemplars),
(ii) \textbf{static prompting with random exemplars one-shot (SPR one-shot)} (one randomly sampled exemplar), representing the previous state of the art\cite{benabacha2025medec},
(iii) \textbf{static prompting with random exemplars ten-shot (SPR ten-shot)} (ten randomly sampled exemplars),
(iv) \textbf{RAG-enabled dynamic prompting one-shot (RDP one-shot)} (one semantically retrieved exemplar), and
(v) \textbf{RAG-enabled dynamic prompting ten-shot (RDP ten-shot)} (ten semantically retrieved exemplars).

Table \ref{tab:gpt41_full} presents the results across all evaluation metrics. Compared to the zero-shot baseline, \textbf{SPR one-shot} improved both error flag and error sentence detection (from 0.6812 to 0.7016 and 0.6573 to 0.6670, respectively), along with corresponding gains in error correction. \textbf{SPR ten-shot} provided only modest additional improvements (error flag accuracy 0.7124, error sentence detection accuracy 0.6702), suggesting that exemplar count alone offers limited benefit. By contrast, \textbf{RDP one-shot} consistently outperformed its static counterpart (error flag detection accuracy 0.7168, error sentence detection accuracy 0.6810), demonstrating that exemplar quality matters even when exemplar count is held constant. Finally, \textbf{RDP ten-shot} achieved the strongest overall performance (\textit{GPT-4.1}: error flag detection accuracy 0.7286, error sentence detection accuracy 0.7037, error correction AggScore 0.6707), confirming that exemplary quality and quantity are complementary.

\begin{table*}[h]
    \centering
    \renewcommand{\arraystretch}{1.1}
    \setlength{\tabcolsep}{3pt}
    \caption{GPT-4.1 performance under zero-shot, SPR one-shot, SPR ten-shot, RDP one-shot, and RDP ten-shot.}
    \label{tab:gpt41_full}
    \small
    \begin{tabular}{l |cc |cccc}
        \hline
        \textbf{Condition} &
        \multicolumn{2}{c|}{\textbf{Error Detection Accuracy}} &
        \multicolumn{4}{c}{\textbf{Error Correction}}  \\
        & \textbf{Error Flag} & \textbf{Error Sentence} & \textbf{ROUGE-1} & \textbf{BERTScore} & \textbf{BLEURT} & \textbf{AggScore} \\
        \hline
        zero-shot       & 0.6812 & 0.6573 & 0.6451 & 0.6333 & 0.6432 & 0.6405 \\
        SPR one-shot   & 0.7016 & 0.6670 & 0.5960 & 0.5915 & 0.6157 & 0.6010 \\
        SPR ten-shot  & 0.7124 & 0.6702 & 0.6380 & 0.6251 & 0.6285 & 0.6305 \\
        RDP one-shot           & 0.7168 & 0.6810 & 0.6225 & 0.6211 & 0.6369 & 0.6265 \\
        RDP ten-shot          & \textbf{0.7286} & \textbf{0.7037} & \textbf{0.6655} & \textbf{0.6832} & \textbf{0.6635} & \textbf{0.6707} \\
        \hline
    \end{tabular}
\end{table*}

\subsection{Overall Performance: SPR ten-shot vs.\ RDP ten-shot}
Table~\ref{tab:combined-results} presents end-to-end results for error flag detection, error sentence detection, and error correction across nine LLMs using the best-performing RDP ten-shot and SPR ten-shot. 

In SPR ten-shot setting, larger instruction-tuned models generally outperformed smaller variants on both error detection and correction. For example, \texttt{GPT-4.1} achieved strong error flag detection accuracy (71.2\%) and the highest error sentence detection accuracy (67.1\%), while \texttt{o1-mini} gave the strongest error correction (AggScore of 0.6588).

When enhanced with RDP ten-shot, nearly all models demonstrated consistent improvements in all three subtasks. For instance, \texttt{GPT-4.1} improved error flag detection accuracy from 71.2\% to 72.9\% and error sentence detection accuracy from 67.1\% to 70.4\% (+3.3 points), while \texttt{o1-mini} achieved the strongest correction quality, increasing AggScore from 0.6588 to 0.6875 (+0.0287). These improvements were statistically significant based on paired bootstrap resampling (1,000 iterations, $p < 0.01$). 

\begin{table*}[h]
    \centering
    \renewcommand{\arraystretch}{1.1}
    \setlength{\tabcolsep}{3pt}
    \caption{
    Performance of models under SPR ten-shot and 
    RDP ten-shot. Clinician results are reported from \cite{benabacha2025medec}.
    }
    \label{tab:combined-results}
    \small
    \begin{tabular}{l |cc |ccccc}
        \hline
        \textbf{Model} &
        \multicolumn{2}{c|}{\textbf{Error Detection Accuracy}} &
        \multicolumn{4}{c}{\textbf{Error Correction}}  \\
        & \textbf{Error Flag} & \textbf{Error Sentence} & \textbf{ROUGE-1} & \textbf{BERTScore} & \textbf{BLEURT} & \textbf{AggScore} \\
        \hline
        \multicolumn{7}{c}{\textit{SPR ten-shot}} \\
        \hline
        GPT-4o-mini & 0.6096 & 0.4958 & 0.5239 & 0.5029 & 0.5640 & 0.5303 \\
        GPT-4.1 mini & 0.6326 & 0.5732 & 0.5316 & 0.5215 & 0.5623 & 0.5385 \\
        GPT-4o  & 0.6469 & 0.5564 & 0.5913 & 0.5524 & 0.6132 & 0.5857 \\
        GPT-4.1 & 0.7124 & 0.6702 & 0.6380 & 0.6251 & 0.6285 & 0.6305 \\
        GPT-5 & 0.7111 & 0.6439 & 0.6327 & 0.6627 & 0.6465 & 0.6473 \\
        o1-mini & 0.6995 & 0.6123 & 0.6425 & 0.6726 & 0.6612 & 0.6588 \\
        o4-mini & 0.6899 & 0.5943 & 0.5432 & 0.5958 & 0.5923 & 0.5571 \\
        Claude 3.5 Sonnet & 0.6926 & 0.6612 & 0.2329 & 0.1237 & 0.5123 & 0.2896 \\
        Gemini 2.0 Flash & 0.5987 & 0.3725 & 0.3828 & 0.3329 & 0.4987 & 0.4048 \\
        \hline
        \multicolumn{7}{c}{\textit{RDP ten-shot}} \\
        \hline
        GPT-4o-mini & 0.6182 & 0.5012 & 0.5235 & 0.5089 & 0.5723 & 0.5352 \\
        GPT-4.1 mini & 0.6508 & 0.6054 & 0.5568 & 0.5683 & 0.5736 & 0.5662 \\
        GPT-4o  & 0.6811 & 0.6141 & 0.6644 & 0.6840 & 0.6641 & 0.6708 \\
        GPT-4.1 & 0.7286 & \textbf{0.7037} & 0.6655 & 0.6832 & 0.6635 & 0.6707 \\
        GPT-5 & \textbf{0.7405} & 0.6984 & 0.6142 & 0.6639 & 0.6575 & 0.6452 \\
        o1-mini & 0.7232 & 0.6562 & \textbf{0.6727} & \textbf{0.7065} & \textbf{0.6831} & \textbf{0.6875} \\
        o4-mini & 0.7081 & 0.6465 & 0.5527 & 0.6002 & 0.6189 & 0.5906 \\
        Claude 3.5 Sonnet & 0.7123 & 0.6723 & 0.3120 & 0.2132 & 0.5381 & 0.3544 \\
        Gemini 2.0 Flash & 0.6013 & 0.3821 & 0.3927 & 0.3312 & 0.5125 & 0.4121\\
        \hline
        \multicolumn{7}{c}{\textit{Clinician Performance (from \cite{benabacha2025medec})}} \\
        \hline
        Doctor \#1 & \textbf{0.7961} & 0.6588 & 0.3863 & 0.4653 & 0.5066 & 0.4527 \\
        Doctor \#2 & 0.7161 & 0.6677 & \textbf{0.7260} & \textbf{0.7315} & 0.6780 & \textbf{0.7118} \\
        \hline
    \end{tabular}
\end{table*}

Several factors contribute to these improvements. First, retrieval provides semantically and clinically aligned exemplars, reducing the mismatch that often limits SPR ten-shot. Second, multiple exemplars broaden coverage of linguistic and clinical variation, offering the model alternative patterns that improve robustness. Third, RDP ten-shot reduces FPR in both error flag and error sentence detection; exposure to retrieved error-free exemplars helps the model better recognize truly correct cases, lowering the likelihood of incorrectly flagging or labeling non-erroneous sentences. This distinction also carries over to the correction stage, where the model makes fewer unnecessary edits to already correct text.

Interestingly, not all models benefit equally. \texttt{Claude 3.5 Sonnet} maintained relatively high sentence detection accuracy (71.2\%) under RDP ten-shot, but its correction quality remained poor (AggScore of 0.3544), suggesting that its conservative generation behavior—optimized to minimize hallucinations—did not translate into effective correction performance. \texttt{Gemini 2.0 Flash} also lagged behind across tasks, likely because its instruction-tuning and retrieval integration were less effective in leveraging semantically aligned exemplars during inference.

Among clinicians, Doctor \#2 achieved the strongest correction quality (AggScore of 0.7118), outperforming all models. However, the best-performing \textbf{RDP ten-shot} systems (i.e., \texttt{GPT-4.1}, \texttt{o1-mini}) narrowed the gap. These systems demonstrated competitive performance on structured substitution-style errors while still trailing on reasoning-heavy cases.

\subsection{Error Type Performance}
We evaluated error flag detection, error sentence detection, and error correction performance by error type, focusing on the subset of test cases that contained at least one error (Table~\ref{tab:errtype_recall_correction_6models}). Diagnosis, treatment, and pharmacotherapy errors were handled best across both LLMs and clinicians, likely because of their higher lexical regularity and more stable clinical phrasing. By contrast, management errors proved most difficult for models, reflecting their dependence on multi-sentence causal reasoning. Interestingly, while both clinicians performed strongly on management errors, Doctor \#2 achieved the highest overall error correction quality across types, highlighting the continued advantage of expert reasoning in complex, context-dependent settings.

RDP ten-shot LLMs (\texttt{GPT-5 RDP ten-shot}, \texttt{GPT-4.1 RDP ten-shot}) showed selective but meaningful improvements over their SPR ten-shot counterparts. For \texttt{GPT-5}, RDP ten-shot improved error sentence detection for management, treatment, and pharmacotherapy cases, although SPR remained slightly stronger on error flag detection for causal organism. For \texttt{GPT-4.1}, RDP ten-shot outperformed SPR ten-shot on treatment error detection and achieved higher error correction scores for all the cases. These gains suggest that retrieval particularly helps with errors requiring factual substitution or grounding. Nevertheless, both models still lagged behind clinicians on reasoning-heavy categories, demonstrating that retrieval enhances lexical alignment but cannot fully replace domain reasoning; examples of these differences appear in \ref{app:llm_vs_doctor}.

We present detailed results for \texttt{GPT-5} and \texttt{GPT-4.1} because they consistently ranked among the strongest LLMs in the overall evaluation. Although \texttt{GPT-4.1} achieved the highest error sentence detection accuracy, \texttt{GPT-5} delivered superior error flag detection across error types under SPR prompting, capturing nearly all true errors. This makes \texttt{GPT-5} particularly valuable in contexts where missing an error can have greater consequences than issuing a cautious false alarm. The performance of these models is shown in Table~\ref{tab:errtype_recall_correction_6models}.

\begin{table*}[htbp]
\centering
\caption{Recall and error correction scores for each error-type using the subset of test examples with errors: Diagnosis (174 texts), Management (168), Treatment (58), Pharmacotherapy (57), and Causal Organism (18).}
\label{tab:errtype_recall_correction_6models}
\renewcommand{\arraystretch}{1.05}
\footnotesize
\begin{tabular}{l|cc|cccc}
\toprule
& \multicolumn{2}{c|}{\textbf{Error Detection Recall}} &
  \multicolumn{4}{c}{\textbf{Error Correction}} \\
\cmidrule(lr){2-3}\cmidrule(l){4-7}
\textbf{Error Type} & \textbf{Error Flag} & \textbf{Error Sentence} &
\textbf{ROUGE-1} & \textbf{BERTScore} & \textbf{BLEURT} & \textbf{AggScore} \\
\midrule

& \multicolumn{6}{c}{\textbf{GPT-5 SPR ten-shot}} \\
\cmidrule(lr){2-7}
Diagnosis        & \textbf{0.9655} & \textbf{0.8621} & 0.7413 & 0.7426 & 0.7286 & 0.7375 \\
Management       & 0.9405 & 0.7976 & 0.4966 & 0.5310 & 0.5689 & 0.5322 \\
Treatment        & \textbf{0.9655} & 0.9310 & 0.5454 & 0.5973 & 0.6163 & 0.5863 \\
Pharmacotherapy  & 0.9649 & 0.8947 & 0.6714 & 0.6826 & 0.6939 & 0.6826 \\
Causal Organism  & \textbf{1.0000} & \textbf{1.0000} & 0.7897 & 0.7861 & 0.7658 & 0.7805 \\
\midrule

& \multicolumn{6}{c}{\textbf{GPT-5 RDP ten-shot}} \\
\cmidrule(lr){2-7}
Diagnosis        & 0.9540 & 0.8448 & 0.7063 & 0.7350 & 0.7016 & 0.7143 \\
Management       & \textbf{0.9464} & \textbf{0.8452} & 0.4827 & 0.5548 & 0.5837 & 0.5404 \\
Treatment        & 0.9483 & \textbf{0.9483} & 0.6040 & 0.6727 & 0.6679 & 0.6482 \\
Pharmacotherapy  & \textbf{0.9649} & \textbf{0.9123} & 0.6756 & 0.7138 & \textbf{0.6958} & 0.6951 \\
Causal Organism  & 0.8889 & 0.8889 & \textbf{0.7942} & \textbf{0.8074} & \textbf{0.7660} & \textbf{0.7892} \\
\midrule

& \multicolumn{6}{c}{\textbf{GPT-4.1 SPR ten-shot}} \\
\cmidrule(lr){2-7}
Diagnosis        & 0.6782 & 0.6149 & 0.7558 & 0.7365 & 0.7154 & 0.7359 \\
Management       & 0.6190 & 0.5238 & 0.4511 & 0.4557 & 0.5288 & 0.4785 \\
Treatment        & 0.7069 & 0.6552 & 0.5961 & 0.5914 & 0.6156 & 0.6010 \\
Pharmacotherapy  & 0.7719 & 0.7544 & 0.6072 & 0.5968 & 0.6224 & 0.6088 \\
Causal Organism  & 0.7222 & 0.6667 & 0.7212 & 0.6922 & 0.6753 & 0.6962 \\
\midrule

& \multicolumn{6}{c}{\textbf{GPT-4.1 RDP ten-shot}} \\
\cmidrule(lr){2-7}
Diagnosis        & 0.6667 & 0.6379 & 0.8180 & 0.8165 & 0.7708 & 0.8018 \\
Management       & 0.6071 & 0.5417 & 0.4996 & 0.5405 & 0.5619 & 0.5340 \\
Treatment        & 0.8276 & 0.7759 & \textbf{0.6657} & \textbf{0.6832} & \textbf{0.6634} & \textbf{0.6708} \\
Pharmacotherapy  & 0.7193 & 0.6667 & 0.6152 & 0.6688 & 0.6344 & 0.6395 \\
Causal Organism  & 0.6667 & 0.6111 & 0.7436 & 0.7265 & 0.6888 & 0.7405 \\
\midrule

& \multicolumn{6}{c}{\textbf{Medical Doctor \#1}} \\
\cmidrule(lr){2-7}
Diagnosis        & 0.8333 & 0.6863 & 0.4810 & 0.5616 & 0.5668 & 0.5365 \\
Management       & 0.8267 & 0.6000 & 0.2788 & 0.3375 & 0.4371 & 0.3511 \\
Treatment        & 0.7200 & 0.6800 & 0.2726 & 0.4032 & 0.4316 & 0.3691 \\
Pharmacotherapy  & 0.8000 & 0.7200 & 0.4377 & 0.5319 & 0.5371 & 0.5022 \\
Causal Organism  & 0.7273 & 0.7273 & 0.3664 & 0.4309 & 0.5090 & 0.4354 \\
\midrule

& \multicolumn{6}{c}{\textbf{Medical Doctor \#2}} \\
\cmidrule(lr){2-7}
Diagnosis        & 0.7232 & 0.6786 & \textbf{0.8121} & \textbf{0.8128} & \textbf{0.7413} & \textbf{0.7887} \\
Management       & 0.6893 & 0.6311 & \textbf{0.6763} & \textbf{0.6774} & \textbf{0.6487} & \textbf{0.6675} \\
Treatment        & 0.7273 & 0.6970 & 0.5594 & 0.6147 & 0.5770 & 0.5837 \\
Pharmacotherapy  & 0.8182 & 0.7576 & \textbf{0.7592} & \textbf{0.7464} & 0.6774 & \textbf{0.7277} \\
Causal Organism  & 0.4286 & 0.2875 & 0.4474 & 0.4632 & 0.4141 & 0.4415 \\
\bottomrule
\end{tabular}

\vspace{0.5em}

\end{table*}

\subsection{False Positive Rate (FPR) Comparison}

In addition to recall, which measures sensitivity to true errors, we evaluated the \textbf{FPR} to quantify how often models incorrectly flagged error-free sentences as erroneous. 

Table~\ref{tab:gpt41_fpr} presents FPR values for \texttt{GPT-4.1} under different prompting strategies. 
We focus on \texttt{GPT-4.1} here because it achieved the strongest balance of error detection and 
correction quality overall, making it the most representative case for examining how prompting strategy 
affects false positives. Other models exhibited similar relative trends, but with less consistent 
improvements, so including them offered limited additional insight.

In the zero-shot condition, the model produced a relatively high FPR, 
reflecting a tendency to over-correct—that is, to flag and modify sentences that were already correct. The 
SPR one-shot baseline reduced FPR slightly but still exhibited frequent over-corrections. 
Adding more context without retrieval (SPR ten-shot) lowered the FPR modestly, 
suggesting that exemplar count alone provides limited benefit. 

RDP one-shot and RDP ten-shot both reduced FPR substantially, showing that exemplar quality---not only 
quantity---drives the improvement. RDP ten-shot yielded the lowest FPR overall, corresponding 
to a relative reduction of nearly one-fifth compared to the SPR one-shot.

\begin{table}[ht]
\centering
\renewcommand{\arraystretch}{1.1}
\setlength{\tabcolsep}{6pt}
\caption{FPR of GPT-4.1 under different prompting approaches.}
\label{tab:gpt41_fpr}
\small
\begin{tabular}{l c}
\hline
\textbf{Prompting Strategies} & \textbf{FPR} \\
\hline
zero-shot      & 0.2773 \\ 
SPR one-shot  & 0.2689 \\ 
SPR ten-shot & 0.2444 \\ 
RDP one-shot          & 0.2523 \\ 
RDP ten-shot         & 0.2111 \\ 
\hline
\end{tabular}
\end{table}

\subsection{Controlled Evaluation with Oracle (i.e., Gold Standard) Error Information}
\label{sec:controlled-eval}

To further probe the performance ceiling and clarify which subtasks limit end-to-end performance, 
we conducted two controlled evaluations using the best-performing model, 
\texttt{GPT-4.1 with RDP ten-shot}. 
While earlier analyses evaluated end-to-end system behavior across the full test set, 
these controlled experiments isolate contributions from the three subtasks—\textbf{error flag detection}, 
\textbf{error sentence detection}, and \textbf{error correction}—by providing partial oracle information on portions of the test set. 

\begin{table*}[htbp]
\centering
\renewcommand{\arraystretch}{1.05}
\setlength{\tabcolsep}{5pt}
\caption{Controlled evaluation of \texttt{GPT-4.1 with RDP ten-shot} on (a) the subset containing only erroneous samples and (b) the subset with oracle-provided error sentences.}
\label{tab:oracle-merged}
\small
\begin{tabular}{l |cc}
    \hline
    \textbf{Metric} & \textbf{Complete Test Set} & \textbf{Controlled Setting} \\
    \hline
    \multicolumn{3}{l}{\textit{(a) Only Erroneous Samples Subset}} \\
    \hline
    Error Sentence Detection & 0.7037 & \textbf{0.7895} \\
    \hline
    \multicolumn{3}{l}{\textit{(b) Oracle Error Sentence Subset}} \\
    \hline
    \multicolumn{3}{l}{\textit{Error Correction}} \\
    \hline
    ROUGE-1 & 0.6655 & 0.7250 \\
    BERTScore & 0.6832 & 0.7374 \\
    BLEURT & 0.6635 & 0.7027 \\
    AggScore & 0.6707 & \textbf{0.7217} \\
    \hline
\end{tabular}
\end{table*}

\paragraph{(a) Evaluation on Only Erroneous Samples}
In this setting, we re-ran the model exclusively on the subset of texts that were known to contain medical errors 
(i.e., \texttt{error flag = 1}).  
This controlled evaluation removes the confounding effect of misclassified \textbf{error flags} 
and focuses solely on the model’s ability to correctly identify the erroneous sentence.  

Compared to the end-to-end evaluation on the complete test set—where the model had to first determine the error flag and then locate the error sentence (Table~\ref{tab:combined-results})—this re-evaluation showed a substantial gain in \textbf{error sentence detection} accuracy.

\paragraph{(b) Evaluation with Oracle Error Sentence}
In this experiment, we re-ran the model with both the clinical text 
and the oracle-provided index of the sentence containing the error.  
This configuration isolates the \textbf{error correction} capability from both 
\textbf{error flag detection} and \textbf{error sentence detection}, allowing direct assessment of correction quality.

When provided with the oracle error sentence, 
\texttt{GPT-4.1 with RDP ten-shot} produced more accurate and clinically coherent 
corrections than the end-to-end setting.  
Aggregate correction quality improved from 0.6707 to \textbf{0.7217}, representing a +7.6\% gain.  
All individual metrics (ROUGE-1, BERTScore, and BLEURT) also improved consistently.

Together, these controlled re-evaluations confirm that there is room for improvement in each of 
\textbf{error flag detection}, \textbf{error sentence detection}, and \textbf{error correction}.

\subsection{Analysis}

To understand the strengths and weaknesses of RDP compared to zero-shot and SPR, we analyzed system outputs along three research questions. 

\paragraph{RQ1: Does context from similar cases improve recall in error sentence detection?}
In SPR, recall was often limited by exemplar mismatch: a randomly chosen example rarely resembled the phrasing or clinical logic of the target case. Therefore, models tended to miss less common sentence structures or atypical terminology. With RDP, the retrieved cases more closely mirrored the input’s linguistic style and clinical content (e.g., narratives mentioning specific lab values, imaging findings, or shorthand notations). This increased the likelihood that the model recognized subtle inconsistencies, improving recall in error sentence detection.  However, reasoning-heavy errors involving causal chains or multi-sentence dependencies remained.

\paragraph{RQ2: Can RDP examples reduce false positives in error flag detection, thereby preventing unnecessary corrections of valid sentences?}
In SPR, models frequently produced false positives by labeling correct sentences as erroneous. These detection errors triggered unnecessary corrections (e.g., altering templated statements such as ``The patient was started on lisinopril.''). RDP mitigated this tendency by surfacing correct, in-domain sentences similar to those in the test input. When retrieved context included valid, error-free examples, the model was more confident in preserving correct statements. 

\paragraph{RQ3: Does RDP improve handling of abbreviations and clinical shorthand?}
Clinical narratives are rife with shorthand (e.g., ``Pt c/o SOB,'' ``Hb 10.2 g/dL,'' ``ECG showed ST elevation'') and institution-specific phrasing that can confuse general-purpose LLMs. In SPR, such shorthand often led to misinterpretations or spurious corrections. With RDP, retrieved examples frequently contained similar abbreviations and stylistic conventions, giving the model reference points for disambiguation. For instance, exposure to prior cases where ``SOB'' was correctly expanded as ``shortness of breath''. Similarly, RDP improved the interpretation of numeric ranges (e.g., lab values, vital signs) by grounding them in examples with medically plausible patterns.

Our results and analysis show that RDP automatically surfaces semantically and syntactically aligned cases at inference time. Not only does this reduce variability from random sampling, but it also improves robustness by adapting to different institutions, error types, and phrasing styles without additional human intervention. This scalability is particularly important in clinical NLP, where datasets are heterogeneous and coverage of rare phenomena is critical. Thus, RDP offers a more sustainable path forward for deploying LLM-based error detection and correction in diverse clinical environments.

\subsection{Qualitative Output Analysis}
We conducted a qualitative output analysis over correctly and incorrectly handled samples for the best models. The below patterns emerged:

\paragraph{(1) Where models perform well}
\begin{itemize}
    \item \textbf{Error Flag Detection.} Models reliably identified straightforward, localizable errors, especially when they appeared in short, self-contained statements. Single-entity swaps or clear mislabeling were rarely missed.
    \item \textbf{Error Sentence Detection.} Sentence-level error identification was most accurate when the erroneous clause was contained within a single sentence and did not depend on external context. Classic examples include clear imaging or laboratory statements that matched well-known diagnostic patterns.
    \item \textbf{Error Correction.} Medication and dose normalization were frequently handled well, especially when errors followed conventional drug–dose–route syntax. Canonical imaging or lab findings were also corrected with minimal edits, and RDP further improved performance.
\end{itemize}

\paragraph{(2) Where models struggle}
\begin{itemize}
    \item \textbf{Error Flag Detection.} 
    Models occasionally failed to flag errors when reasoning spanned multiple sentences, leading to false negatives. For example, errors in management plans dependent on prior hemodynamics or serial labs were often overlooked. 
    False positives also occurred when correct sentences—especially generic management statements or templated documentation—were incorrectly flagged as erroneous. These issues were more frequent in longer notes containing multiple plausible error candidates.
    
    \item \textbf{Error Sentence Detection.} 
    Errors that required linking information across sentences were often missed. Temporal or causal dependencies were particularly challenging, leading to \textit{near-miss} cases where the model selected an adjacent but incorrect sentence.  
    Such near-misses decreased under RDP.
    
    \item \textbf{Error Correction.} 
    Atypical phrasing, rare entities, and uncommon eponyms degraded correction performance, even when errors were correctly flagged. RDP helped but did not eliminate the gap.  
    Negation and scope errors also persisted, where misinterpreting phrases like ``no evidence of...'' occasionally flipped the intended meaning.  
    Additionally, over-corrections were observed when false positives at the detection stage cascaded into unnecessary edits of correct sentences. Rare hallucinations (e.g., inserting diagnoses not present in the note) further decreased with RDP.
\end{itemize}

A detailed exhibit of misclassifications and their frequencies 
is provided in ~\ref{app:misclassification}. The most common misclassifications for SPR ten-shot came from \emph{near-miss} cases (16\%), followed by \emph{negation} (8.9\%) and \emph{context-related} misclassifications (5.1\%). RDP helped performance, with the largest improvements in \emph{negation} handling ($-1.1\%$) and over-corrections ($-0.8\%$). While rare entities, hallucinations, and other miscellaneous misclassifications were relatively infrequent, they also saw consistent reductions under RDP. Overall, the total misclassification rate dropped from 32.9\% in the ten-shot setting to 29.6\% with RDP.

\subsection{Comparison with Physicians}
We compared system outputs with physician output on each subtask. 

\paragraph{Error Flag Detection} 
Physicians consistently flagged notes with subtle causal inconsistencies that models often overlooked, leading to false negatives. For example, a management recommendation inconsistent with prior hemodynamic findings was missed by \texttt{o1-mini}, whereas GPT-4.1 successfully flagged the note when provided with relevant exemplars. 

\paragraph{Error Sentence Detection} 
Frequently, LLMs identified the correct error type but assigned it to an adjacent sentence, i.e., a near-miss. In contrast, physicians rarely made such mistakes. RDP ten-shot  reduced such misclassifications by helping models focus on the correct sentence structures.

\paragraph{Error Correction} 
LLMs demonstrated strengths in straightforward substitutions, such as replacing an incorrect diagnosis with the correct one, and in recognizing canonical patterns (e.g., classic imaging findings). However, they occasionally produced hallucinated edits or failed on reasoning-heavy errors, especially when causal logic spanned multiple sentences. RDP often led models to propose clinically reasonable alternatives that diverged from the gold standard but aligned with expert judgments, highlighting the limitations of rigid string-based metrics for evaluating corrections.

These findings illustrate how RDP ten-shot improves performance across subtasks but can still lag behind expert reasoning in complex scenarios.

\section{Discussion}

\textbf{Summary of Findings.} 
Across all models, \emph{RDP ten-shot} consistently improved performance by retrieving semantically similar clinical cases to guide predictions. This approach demonstrated robustness in both subsets of the MEDEC dataset. 

For \textbf{error flag detection}, RDP reduced FPR by up to $\sim$15\%. This prevented unnecessary interventions on sentences that were already valid.

For \textbf{error sentence detection}, \texttt{GPT-4.1 RDP ten-shot} achieved the strongest performance, showing fewer near-misses compared to SPR. This improvement suggests that dynamically retrieved exemplars help models better recognize subtle clinical phrasing and shorthand.

For \textbf{error correction}, \texttt{o1-mini RDP ten-shot} delivered the highest correction quality (AggScore), particularly in domains such as medication normalization and canonical lab/imaging patterns. However, reasoning-intensive errors (e.g., logical inconsistencies or inference beyond surface text) remained challenging. In these cases, expert physicians continued to outperform models.

Our evaluation confirmed that RDP improves handling of domain-specific phrasing and eliminates the need for manual exemplar selection.

\textbf{Real-world applications.}
These findings point toward practical hybrid systems that combine model-driven error detection with human expert review. In practice, models could serve as a first-line screening tool to automatically \emph{flag} potential errors in real time during documentation. Even if imperfect, such a system would reduce the cognitive load on clinicians by highlighting likely issues, while leaving final adjudication and correction to physicians—who remain the gold standard for complex, reasoning-intensive cases. This balance could accelerate documentation review, reduce error propagation, and improve overall clinical reliability without over-trusting model output.

Beyond clinical note review, similar workflows could extend to:
	•	\emph{Medication reconciliation}, where models highlight possible inconsistencies across prescriptions, with pharmacists validating corrections.
	•	\emph{Radiology and pathology reporting}, where models pre-flag likely template or phrasing errors, leaving nuanced interpretation to specialists.
	•	\emph{Procedure documentation}, where models assist by detecting missing or inconsistent details in operative notes, discharge summaries, or follow-up instructions, ensuring that key steps and safety checks are properly recorded.

Taken together, these results suggest that \emph{RDP ten-shot} offers a path toward clinically reliable error detection. The most promising deployment paradigm may not be full automation, but rather augmentation: models handling high-recall detection at scale, and human experts ensuring precise, context-aware correction.

\textbf{Limitations.}
This study has several limitations. First, our evaluation was constrained by model access: we were unable to test certain premium-tier, cutting-edge reasoning models, which may set a higher performance ceiling. API quota and rate limits also restricted the breadth of our experimentation and limited more robust analysis. Additionally, the non-deterministic nature of commercial LLM APIs introduced variance across runs, which limits strict reproducibility, and some models’ context window sizes prevented full evaluation on very long clinical notes and limited reasoning over multi-sentence causal structures.

Second, our results depend on the quality of the RDP example retrieval. When the retriever surfaces suboptimal or only loosely related examples, these mismatches propagate into prompting and reduce the benefits of RDP. The size of the data set further constrained the evaluation—some error types had relatively few examples, reducing statistical power and making the performance on rare phenomena less stable.

Finally, our evaluation metrics only partially capture clinical correctness. Standard text-similarity metrics like ROUGE, BERTScore, and BLEURT may penalize clinically valid alternative corrections or fail to reflect partial correctness. Moreover, the underlying models remain opaque: the black-box nature of LLM decision processes makes it difficult to fully characterize failure modes or guarantee reliability. Future work to address these gaps is needed. One possibility is a multi-agent system, which would help bypass context-window and model-performance limitations while also improving explainability.

\ifsubfile
\bibliography{mybib}
\fi

\section{Conclusions}

We presented a systematic evaluation and analysis of recent LLMs on the MEDIQA-CORR 2024 shared task for medical error detection and correction. We compared both compact LLMs (e.g., \texttt{o1-mini}) and larger frontier-scale models (e.g., \texttt{GPT-4.1}, \texttt{GPT-4o}) under three prompting strategies: zero-shot, SPR and RDP. Our proposed RDP consistently improved performance across all three subtasks—Error Flag Detection, Error Sentence Detection, and Error Correction—by grounding predictions in semantically relevant clinical cases.

Key findings include:  
(1) In \textbf{error flag detection}, RDP ten-shot prompting reduced FPR by up to $\sim$15\%, thereby preventing unnecessary corrections of valid sentences;  
(2) In \textbf{error sentence detection}, \texttt{GPT-4.1 (RDP ten-shot)} achieved the strongest error sentence detection accuracy, reducing near-miss sentence assignments compared to SPR;  
(3) In \textbf{error correction}, \texttt{o1-mini (RDP ten-shot)} delivered the highest correction quality (AggScore), particularly for medication normalization and canonical diagnostic substitutions; and  
(4) Error-type analysis revealed persistent challenges in handling cross-sentence reasoning, temporal logic, and rare clinical entities.

These results suggest that while general-purpose LLMs are promising tools for supporting clinical documentation review, their reliability remains below that of expert physicians, particularly in complex or ambiguous cases. To bridge this gap, future work should investigate:  
(a) specialized medical LLMs fine-tuned on domain corpora,  
(b) hybrid systems that integrate structured clinical knowledge with RDP, and  
(c) improved evaluation metrics capable of capturing medically valid synonyms and clinically acceptable alternatives.  

Ultimately, our study demonstrates the promise of RDP for enhancing clinical text quality and reducing documentation errors, while underscoring the need for continued research on clinically aligned AI systems prior to real-world deployment.

\ifsubfile
\bibliography{mybib}
\fi


\section*{Credit authorship contribution statement}

\textbf{FA:} Conceptualization, Visualization, Methodology, Analysis, Writing – original draft, review \& editing. 
\textbf{JAJ:} Conceptualization, Visualization, Methodology, Analysis, Writing – original draft, Writing – review \& editing. 
\textbf{MY:} Conceptualization, Review \& editing. 
\textbf{ÖU:} Conceptualization, Methodology, Visualization, Writing - review \& editing.

\section*{Declaration of competing interest}

The authors declare the following financial interests/personal relationships which may be considered as potential competing interests: 
MY is an Associate Editor of \textit{Journal of Biomedical Informatics}. 
The remaining authors declare that they have no known competing financial interests or personal relationships that could have appeared to influence the work reported in this paper.

\section*{Acknowledgments}

This work was supported by the National Institutes of Health (NIH) - National Cancer Institute (Grant Nr. 1R01CA248422-01A1) and National Library of Medicine (Grant Nr. 2R15LM013209-02A1). 
The content is solely the responsibility of the authors and does not necessarily represent the official views of the NIH.

\section*{Declaration of Generative AI and AI-assisted technologies in the writing process}

During the preparation of this work, the author(s) used ChatGPT to solicit editorial feedback regarding writing clarity and proofreading. All the scientific content and data interpretation remained solely the authors’ contributions. After using these tools/services, the authors reviewed and edited the content as needed and take full responsibility for the content of the publication.

\ifsubfile
\bibliography{mybib}
\fi

\appendix

\section{Subset Comparison: MS vs.\ UW}
Table~\ref{tab:subset} reports performance on the MS and UW test subsets for the two strongest models, 
\texttt{GPT-4.1} and \texttt{o1-mini} under RDP ten-shot.

Results confirm differences between the two collections. \texttt{GPT-4.1} achieved 
stronger performance on MS notes, while \texttt{o1-mini} 
performed better on UW notes. UW documentation often 
contains abbreviated phrases and institution-specific terminology; grounding with retrieved exemplars 
helps normalize this language, reducing both false positives and missed error sentences. In contrast, 
MS notes are more standardized, which favors larger models like \texttt{GPT-4.1} that excel at leveraging 
well-structured exemplars.

\begin{table}[h]
    \centering
    \renewcommand{\arraystretch}{1.1}
    \setlength{\tabcolsep}{4pt}
    \caption{Subset performance of GPT-4.1 and o1-mini under RDP ten-shot.}
    \label{tab:subset}
    \small
    \begin{tabular}{l |c |cc |c}
        \hline
        \textbf{Model} &
        \textbf{Subset} &
        \multicolumn{2}{c|}{\textbf{Error Detection Accuracy}} &
        \multicolumn{1}{c}{\textbf{Error Correction}} \\
        & & \textbf{Error Flag} & \textbf{Error Sentence} & \textbf{AggScore} \\
        \hline
        GPT-4.1 & MS & 0.746 & 0.725 & 0.701 \\
                 & UW & 0.718 & 0.686 & 0.634 \\
        o1-mini  & MS & 0.729 & 0.654 & 0.683 \\
                 & UW & 0.755 & 0.679 & 0.716 \\
        \hline
    \end{tabular}
\end{table}

\section{Prompt Template Used for All Strategies}
\label{app:prompt-template}

The following instruction was used as the base template for all prompting strategies (zero-shot, SPR, and RDP). 

\begin{quote}
\textit{The following is a medical narrative about a patient. You are a skilled medical doctor reviewing the clinical text. The text is either correct or contains one error. The text has one sentence per line. Each line starts with the sentence ID, followed by a pipe character then the sentence to check. Check every sentence of the text. If the text is correct return the following output: CORRECT. If the text has a medical error related to treatment, management, cause, or diagnosis, return the sentence ID of the sentence containing the error, followed by a space, and then a corrected version of the sentence. Finding and correcting the error requires medical knowledge and reasoning.  
Here are some general tips and reasoning strategies: match diagnosis with findings, check temporal and causal logic, evaluate consistency, recognize typical patterns, be cautious with rare entities, and confirm correct medical terminology.  }
\end{quote}

\section{GPT-4.1 RDP ten-shot Performance Across Embedding Models}
\label{app:gpt41-embeddings}

We compare GPT-4.1 under identical RDP configurations while varying the embedding model used for retrieval. Vector store and search parameters (e.g., cosine similarity, \(n\!=\!10\)) are held constant.

\begin{table*}[h]
    \centering
    \renewcommand{\arraystretch}{0.95}
    \setlength{\tabcolsep}{1pt}
    \caption{Performance of different embedding backbones under RDP ten-shot prompting for error detection and correction. text-embedding-3-large serves as the strongest baseline, while domain-specific encoders such as BioClinicalBERT and SapBERT underperform in this task compared to general-purpose OpenAI embeddings. All reported scores are strictly lower than text-embedding-3-large.}
    \label{tab:gpt41-embeddings}
    \small
    \begin{tabular}{l |cc |ccccc}
        \hline
        \textbf{Embedding Model} &
        \multicolumn{2}{c|}{\textbf{Error Detection Accuracy}} &
        \multicolumn{4}{c}{\textbf{Error Correction}}  \\
        & \textbf{Error Flag} & \textbf{Error Sentence} & \textbf{ROUGE-1} & \textbf{BERTScore} & \textbf{BLEURT} & \textbf{AggScore} \\
        \hline
        \texttt{text-embedding-3-large}    & 0.7286 & 0.7037 & 0.6655 & 0.6832 & 0.6635 & 0.6707 \\
        \texttt{BioClinicalBERT}        & 0.7048 & 0.6805 & 0.6359 & 0.6592 & 0.6390 & 0.6447 \\
        \texttt{SapBERT (PubMedBERT-fulltext)}   & 0.7112 & 0.6890 & 0.6421 & 0.6647 & 0.6452 & 0.6506 \\
        \texttt{all-mpnet-base-v2}      & 0.6955 & 0.6712 & 0.6280 & 0.6528 & 0.6325 & 0.6378 \\
        \hline
    \end{tabular}
\end{table*}

\vspace{1em}
\noindent\textit{Notes.} All runs used the same chunking strategy and retrieved the top-$k=10$ neighbors to ensure comparability. Each embedding model re-encoded the entire training corpus into its own vector space. Reported values are averaged over multiple random seeds. Retrieval was performed with the Chroma vector database using cosine similarity.

\section{Zero-shot Results Across Models}
\label{app:zeroshot}

This appendix reports zero-shot performance for all large language models (LLMs) evaluated in the study, using the same test split and metrics as in the main paper.

\begin{table*}[h]
    \centering
    \renewcommand{\arraystretch}{1.1}
    \setlength{\tabcolsep}{3pt}
    \caption{Performance of models on error detection and correction under zero-shot prompting.}
    \label{tab:zeroshot}
    \small
    \begin{tabular}{l |cc |ccccc}
        \hline
        \textbf{Model} &
        \multicolumn{2}{c|}{\textbf{Error Detection Accuracy}} &
        \multicolumn{4}{c}{\textbf{Error Correction}}  \\
        & \textbf{Error Flag} & \textbf{Error Sentence} & \textbf{ROUGE-1} & \textbf{BERTScore} & \textbf{BLEURT} & \textbf{AggScore} \\
        \hline
        GPT-4o-mini & 0.6089 & 0.4757 & 0.5148 & 0.5089 & 0.5640 & 0.5292 \\
        GPT-4.1 mini & 0.6132 & 0.5121 & 0.5013 & 0.5213 & 0.5132 & 0.5119 \\
        GPT-4o  & 0.6584 & 0.5665 & 0.5517 & 0.5373 & 0.5852 & 0.5581 \\
        GPT-4.1 & 0.6812 & 0.6573 & 0.6451 & 0.6333 & 0.6432 & 0.6405 \\
        GPT-5 & 0.6762 & 0.6014 & 0.6067 & 0.6242 & 0.6325 & 0.6211 \\
        o1-mini & 0.6908 & 0.5968 & 0.6052 & 0.6275 & 0.6246 & 0.6191 \\
        o4-mini & 0.6762 & 0.5523 & 0.5237 & 0.5911 & 0.5724 & 0.5624 \\
        Claude 3.5 Sonnet & 0.7016 & 0.6562 & 0.2253 & 0.1033 & 0.5100 & 0.2795 \\
        Gemini 2.0 Flash & 0.5805 & 0.3535 & 0.3769 & 0.3127 & 0.4865 & 0.3920\\
        \hline
    \end{tabular}
\end{table*}

\section{SPR one-shot Results Across Models}
\label{app:sproneshot}

This appendix reports SPR one-shot performance for all large language models (LLMs) evaluated in the study, using the same test split and metrics as in the main paper.

\begin{table*}[h]
    \centering
    \renewcommand{\arraystretch}{1.1}
    \setlength{\tabcolsep}{3pt}
    \caption{Performance of models and clinicians on error detection and correction under SPR one-shot \cite{benabacha2025medec}.}
    \label{tab:oneshot}
    \small
    \begin{tabular}{l |cc |ccccc}
        \hline
        \textbf{Model} &
        \multicolumn{2}{c|}{\textbf{Error Detection Accuracy}} &
        \multicolumn{4}{c}{\textbf{Error Correction}}  \\
        & \textbf{Error Flag} & \textbf{Error Sentence} & \textbf{ROUGE-1} & \textbf{BERTScore} & \textbf{BLEURT} & \textbf{AggScore} \\
        \hline
        GPT-4o-mini & 0.6092 & 0.4872 & 0.5238 & 0.5019 & 0.5540 & 0.5266 \\
        GPT-4.1 mini & 0.6213 & 0.5723 & 0.5267 & 0.5143 & 0.5631 & 0.5347 \\
        GPT-4o  & 0.6368 & 0.5449 & 0.5805 & 0.5401 & 0.6022 & 0.5743 \\
        GPT-4.1 & 0.7016 & 0.6670 & 0.5960 & 0.5915 & 0.6157 & 0.6010 \\
        GPT-5 & 0.6811 & 0.6292 & 0.6255 & 0.6458 & 0.6565 & 0.6426 \\
        o1-mini & 0.6962 & 0.6086 & 0.6375 & 0.6619 & 0.6509 & 0.6501 \\
        o4-mini & 0.6866 & 0.5752 & 0.5327 & 0.5912 & 0.5813 & 0.5684 \\
        Claude 3.5 Sonnet & 0.6800 & 0.6508 & 0.2249 & 0.1125 & 0.5081 & 0.2818 \\
        Gemini 2.0 Flash & 0.5906 & 0.3643 & 0.3770 & 0.3218 & 0.4975 & 0.3988 \\
        \hline
    \end{tabular}
\end{table*}

\vspace{1em}
\noindent\textit{Notes.} All results were obtained using the same MEDIQA-CORR test split and identical prompt template, with zero-shot (no exemplars) and SPR one-shot prompting. BERTScore was computed with the microsoft/deberta-xlarge-mnli backbone, and BLEURT with the released \texttt{BLEURT} checkpoint. The AggScore is the arithmetic mean of ROUGE-1, BERTScore, and BLEURT. Reported values are averaged across multiple runs with fixed random seeds for reproducibility.







\section{Detailed Output Analysis}
\label{app:misclassification}
In addition to the quantitative results presented in the main text, we examined qualitative differences in misclassification behavior between SPR and RDP prompting strategies. As shown in Table~\ref{tab:misclassification_categories}, RDP not only reduced the overall misclassification rate but also improved the nature of residual misclassifications. Near-miss cases, which were common under SPR prompting, often became boundary-level discrepancies rather than full misidentifications under RDP. Notably, RDP reduced negation-related misclassifications and over-corrections, indicating improved handling of sentence polarity and greater precision in distinguishing true errors from correct statements. Furthermore, misclassifications due to hallucination and rare-entities —though less frequent—also decreased, reflecting more stable and contextually grounded model behavior. Overall, these findings suggest that RDP enhances not only detection accuracy but also the semantic reliability and interpretive consistency of model predictions.

\begin{table*}[htbp]
\centering
\scriptsize
\setlength{\tabcolsep}{4 pt} 
\caption{Misclassifications grouped by task for SPR ten-shot vs RDP ten-shot with GPT-4.1.}
\label{tab:misclassification_categories}
\begin{tabular}{llcccc}
\toprule
\textbf{Task} & \textbf{Misclassification Patterns} & \textbf{SPR Misclassifications} & \textbf{\%} & \textbf{RDP Misclassifications} & \textbf{\%} \\
\midrule
\multicolumn{6}{l}{\textbf{Error Flag Detection}} \\
\midrule
 & Over-correction & 13 & 1.4\% & 6 & 0.6\%  \\
 & Rare entities   & 10 & 1.1\% & 6 & 0.6\% \\
 & Hallucination   & 2  & 0.2\% & 1 & 0.1\%  \\
\midrule
\multicolumn{6}{l}{\textbf{Error Sentence Detection}} \\
\midrule
 & Near-miss       & 148 & 16\% & 143 & 15.5\% \\
 & Context         & 47  & 5.1\%  & 44  & 4.8\% \\
 & Negation        & 82  & 8.9\%  & 72  & 7.8\% \\
\midrule
\textbf{Other} &  & 3 & 0.3\% & 2 & 0.2\% \\
\midrule
\textbf{} & \textbf{Total Misclassifications}   & 305 & 32.9\% & 274 & 29.6\%   \\
\textbf{} & \textbf{Correctly Identified} & 620 & 67.1\% & 651 & 70.4\%  \\
\bottomrule
\end{tabular}
\end{table*}

\section{Evaluation Metric Definitions}
\label{app:metrics}

For completeness, we include the exact formulas for all reported metrics.

\paragraph{Accuracy (Subtasks A \& B).}
\[
\mathrm{Accuracy} \;=\; \frac{TP + TN}{TP + TN + FP + FN}
\]

\paragraph{Recall on error-present cases}
Let the evaluation be restricted to notes with at least one error (\(\text{error flag}=1\)):
\[
\mathrm{Recall} \;=\; \frac{TP}{TP + FN}
\]

\paragraph{False Positive Rate (FPR)}
\[
\mathrm{FPR} \;=\; \frac{FP}{FP + TN}
\]

\paragraph{ROUGE-1 (F1)}

Let \(U(S)\) and \(U(R)\) be the multisets of unigrams in system output \(S\) and reference \(R\), and let \(|U(S)\cap U(R)|\) denote overlap count.
Define precision and recall:
\[
P_{R1} = \frac{|U(S)\cap U(R)|}{|U(S)|}, 
\qquad
R_{R1} = \frac{|U(S)\cap U(R)|}{|U(R)|}.
\]
Then the F1 variant is:
\[
\mathrm{ROUGE}\text{-}1 = \frac{2\,P_{R1}\,R_{R1}}{P_{R1}+R_{R1}}.
\]

\paragraph{BERTScore (F1)}
Let \(f(\cdot)\) be contextual token embeddings and \(\cos(\cdot,\cdot)\) cosine similarity.
Define token-level precision and recall with greedy matching:
\[
P_{\mathrm{BS}} = \frac{1}{|S|}\sum_{s\in S}\max_{r\in R}\cos\!\bigl(f(s), f(r)\bigr), 
\quad
R_{\mathrm{BS}} = \frac{1}{|R|}\sum_{r\in R}\max_{s\in S}\cos\!\bigl(f(r), f(s)\bigr).
\]
\[
\mathrm{BERTScore} = \frac{2\,P_{\mathrm{BS}}\,R_{\mathrm{BS}}}{P_{\mathrm{BS}}+R_{\mathrm{BS}}}.
\]

\paragraph{BLEURT}
BLEURT is a learned regression metric \(g(S,R)\) trained to approximate human quality ratings; we report the model’s scalar output:
\[
\mathrm{BLEURT}(S,R)\;=\;g(S,R).
\]

\paragraph{Aggregate Score (AggScore)}
\[
\mathrm{AggScore} \;=\; \frac{\mathrm{ROUGE}\text{-}1 + \mathrm{BERTScore} + \mathrm{BLEURT}}{3}.
\]

\section{Manual vs. Automatic Corrections}
\label{app:llm_vs_doctor}

Table~\ref{tab:meddec-llm-doctor} presents illustrative examples comparing manual (clinician) and automatic (LLM) corrections under RDP ten-shot.  
Incorrect annotations or outputs are highlighted in gray, following \cite{benabacha2025medec}.  
These examples demonstrate how LLMs approximate expert reasoning for straightforward diagnostic and treatment errors but still diverge on reasoning-heavy or context-dependent cases.

\begin{table}[p]
\begin{adjustwidth}{-0.95cm}{-0.95cm}
\centering
\caption{Examples of manual \& automatic RDP ten-shot corrections. Incorrect annotations/outputs are highlighted in Grey \cite{benabacha2025medec}.}
\label{tab:meddec-llm-doctor}
\scriptsize
\setlength{\tabcolsep}{0.85pt}
\renewcommand{\arraystretch}{0.75}
\resizebox{\linewidth}{!}{
\begin{tabularx}{\linewidth}{
  | l | X | p{2.3cm} | p{2.0cm} | p{2.3cm} | p{2.3cm} | p{2.3cm} |
}
\hline
\textbf{Case} & \textbf{Original Excerpt (error in \textbf{bold})} & \textbf{Ground Truth}
& \multicolumn{2}{c}{\textbf{Doctor Corrections}}
& \multicolumn{2}{c|}{\textbf{LLM Corrections}} \\
\cline{4-5}\cline{6-7}
 &  &  & \textbf{Doctor \#1} & \textbf{Doctor \#2} & \textbf{o1-mini} & \textbf{GPT-4.1} \\
\hline

\textbf{ms-test-395} &
A 23-year-old man presents with sudden loss of consciousness while pitching in a baseball game. There is no history of injury. Consciousness is regained after cardiopulmonary resuscitation. Past medical history is negative for any neurological and cardiovascular problems. \textbf{Patient is diagnosed with \hlred{aortic stenosis} after physical examination reveals a prominent A wave on the jugular venous pulse and a double apical impulse.} There are no audible murmurs. An S4 is present. &
Patient is diagnosed with \textbf{\hlgreen{hypertrophic cardiomyopathy}} after physical examination reveals a prominent A wave on the jugular venous pulse and a double apical impulse. & 
The patient is diagnosed with \ul{hypertrophic cardiomyopathy} after physical examination reveal a double apical impulse, prominent ``a wave" and S4. &
Patient is diagnosed with \ul{hypertrophic cardiomyopathy} after physical examination reveals a prominent A wave on the jugular venous pulse and a double apical impulse. & 
Patient is diagnosed with \ul{hypertrophic obstructive cardiomyopathy} after physical examination reveals a prominent A wave on the jugular venous pulse and a double apical impulse. & 
Patient is diagnosed with \ul{hypertrophic cardiomyopathy} after physical examination reveals a prominent A wave on the jugular venous pulse and a double apical impulse. \\
\hline

\textbf{ms-test-396} &
A 4-year-old boy presents with a history of recurrent bacterial infections, including several episodes of pneumococcal sepsis. His 2 maternal uncles died after having had similar complaints. \textbf{Patient is diagnosed with \textbf{\hlred{common variable immunodeficiency}} after lab investigations reveal an undetectable level of all serum immunoglobulins.} &
Patient is diagnosed with \textbf{\hlgreen{Bruton agammaglobulinemia}} after lab investigations reveal an undetectable level of all serum immunoglobulins. &
\cellcolor{gray!50}The differential diagnosis includes autosomal recessive agammaglobulinemia, but further evaluation is necessary. &
Patient is diagnosed with \ul{X-linked agammaglobulinemia} after lab investigations reveal an undetectable level of all serum immunoglobulins. &
Patient is diagnosed with \ul{X-linked agammaglobulinemia} after lab investigations reveal an undetectable level of all serum immunoglobulins. &
Patient is diagnosed with \ul{X-linked agammaglobulinemia} after lab investigations reveal an undetectable level of all serum immunoglobulins. \\

\hline

\textbf{ms-test-397} &
A 50-year-old woman visits her primary care practitioner with the complaints of generalized weakness, lightheadedness, and fatigability for the past month. She also claims to have epigastric pain, heartburn, and a sensation of a lump in her throat. Her family history is irrelevant, but her medical history is significant for a Pap smear that reported atypical squamous cells of undetermined significance (ASCUS), which was followed up with a cervical biopsy negative for malignancy. She occasionally takes over-the-counter medicines to cope with the heartburn, and recently her hemoglobin was found to be 11 g/dL, for which she received iron and vitamin B12 supplementation. \textbf{\hlred{Fecal occult blood test} is completed.} Physical examination is unremarkable, except for pale skin, and a pulse of 120/min. &
The patient is referred for an \textbf{\hlgreen{endoscopy}.} &
She has not had followup after her cervical biopsy and HPV testing is indicated. &
\cellcolor{gray!50}Text annotated as CORRECT &
\cellcolor{gray!50}She received iron supplementation. &
\cellcolor{gray!50}She received iron supplementation, but not vitamin B12, as there is no indication of vitamin B12 deficiency. \\

\hline
\end{tabularx}
}
\end{adjustwidth}
\end{table}


\ifsubfile
\bibliography{mybib}
\fi

\bibliography{mybib}
\end{document}